\documentclass[runningheads]{llncs}
\usepackage[T1]{fontenc}
\usepackage{multirow}
\usepackage{graphicx}
\usepackage{booktabs}
\begin{document}
\title{Democratizing AI in Africa: Federated Learning for Low-Resource Edge Devices}
\titlerunning{Democratizing AI in Africa: FL for Low-Resource Edge Devices}
%
 \author{
 Jorge Fabila \inst{1}\and
 Víctor M. Campello \inst{1}\and
 Carlos Martín-Isla \inst{1}\and
 Johnes Obungoloch \inst{2}\and
 Kinyera Leo \inst{2} \and
 Amodoi Ronald \inst{2}
 Karim Lekadir \inst{1,3}
 }
%

\authorrunning{Fabila et al.}
%
 \institute{
 $^1$Dept. de Matemàtiques i Informàtica, Universitat de Barcelona, Barcelona, Spain\\
 $^2$Mbarara University of Science and Technology, Uganda\\
 $^3$Institució Catalana de Recerca i Estudis Avançats (ICREA), Barcelona, Spain
 }

%
\maketitle              
\begin{abstract}

Africa faces significant challenges in healthcare delivery due to limited infrastructure and access to advanced medical technologies. This study explores the use of federated learning to overcome these barriers, focusing on perinatal health. We trained a fetal plane classifier using perinatal data from five African countries: Algeria, Ghana, Egypt, Malawi, and Uganda, along with data from Spanish hospitals. To incorporate the lack of computational resources in the analysis, we considered a heterogeneous set of devices, including a Raspberry Pi and several laptops, for model training. We demonstrate comparative performance between a centralized and a federated model, despite the compute limitations, and a significant improvement in model generalizability when compared to models trained only locally. These results show the potential for a future implementation at a large scale of a federated learning platform to bridge the accessibility gap and improve model generalizability with very little requirements.


\end{abstract}
\section{Introduction}

Today's advancements in Artificial Intelligence (AI) are proposed at a blistering pace and rely greatly on the availability of large-scale and varied datasets. As a result, the training of AI models is becoming restricted to users with access to such datasets and a high-performance computing cluster with a large storage that can train a model in a reasonable amount of time.

These requirements might threaten the democratization of AI tools, and specifically, its use on low-resource settings, worsening the accessibility gap [13]. One particular region where reducing this gap is paramount is Africa, a continent that faces deep challenges such as widespread poverty, limited infrastructure, and insufficient access to quality medical services. In fact, AI presents a unique opportunity to improve healthcare access in Africa. For instance, by providing support to healthcare workers with acquiring medical images or extracting biomarkers or risk scores in the absence of the appropriate imaging modality or clinical expert. Several studies have assessed the viability of such an approach, for example, for oral cancer screening in the absence of clinical experts in India [14] or for simplifying image acquisition of fetal ultrasound and estimating gestational age from non-standard planes [15].

Federated learning (FL) has been proposed to help circumvent the requirement of having access to a large database [16]. With this technique, the desired model design and weights are shared across a set of nodes in each participating institution and trained with a corresponding local dataset. When finished, the model weights are sent back to a central node that combines the different weights received to maximize final performance. This process is usually repeated several times until the improvements stall. By following this approach one avoids transferring the different datasets to one central node, guaranteeing a stronger data privacy. Several works have tested this learning technique in simulated scenarios [2][3][4] and recently, one study conducted a real-world validation with Raspberry Pi devices in three institutions in the United Kingdom [8], but only using classical machine learning. Another study trained a federated model across 20 clients using COVID-19 cases [19], although no details were provided about the infrastructure and all nodes are assumed to have a GPU. In all cases, the different nodes had either the same hardware devices or had a GPU. To our knowledge, no existing work has considered a real heterogeneous setting where devices with different capabilities are involved in the training process.

In this study, we aim to develop the first scientific framework for inclusive imaging AI in resource-limited settings bringing the current state-of-the-art methods (mostly developed for high-income settings) towards new imaging AI algorithms that are fundamentally inclusive, i.e. affordable for resource-limited clinical centres, scalable to under-represented population groups, and accessible to minimally trained clinical workers. With this we aim to bridge the healthcare accessibility gap, providing essential support for working with medical imaging to those who need it most.

In this work, we focus on the enhancement of obstetric ultrasound screening in rural Africa, with the aim of empowering midwives with AI-based tools through the implementation of an FL platform that aims to overcome the hardware constrains in some healthcare facilities. With our FL platform we enable collaboration across borders by training a fetal ultrasound plane classifier with data from Algeria, Ghana, Egypt, Malawi, and Uganda. We demonstrate the added value of collective training for creating robust and generalizable AI models even when working with edge devices, such as a Raspberry Pi device.

\section{Methodology}

\subsection{Datasets}
For this study, we used two publicly available fetal ultrasound datasets which consisted of standard fetal ultrasound planes acquired during routine clinical practice on the 2nd and 3rd trimester of pregnancy.

On one hand, we considered a large-scale dataset consisting of 12,000 images acquired at Hospital Clínic and Hospital Sant Joan de Déu in Barcelona, Spain [7]. The images were acquired with several curvilinear transducers: Voluson E6, Voluson S8 and Voluson S10 (GE Medical Systems, Zipf, Austria), and Aloka (Aloka CO., LTD.)

On the other hand, we considered a small-scale dataset from five African countries (Algeria, Egypt, Ghana, Malawi and Uganda) containing a total of 450 images. These images were acquired with a diversity of ultrasound devices. The specific details can be found in [5].

\begin{figure}
\includegraphics[width=\textwidth]{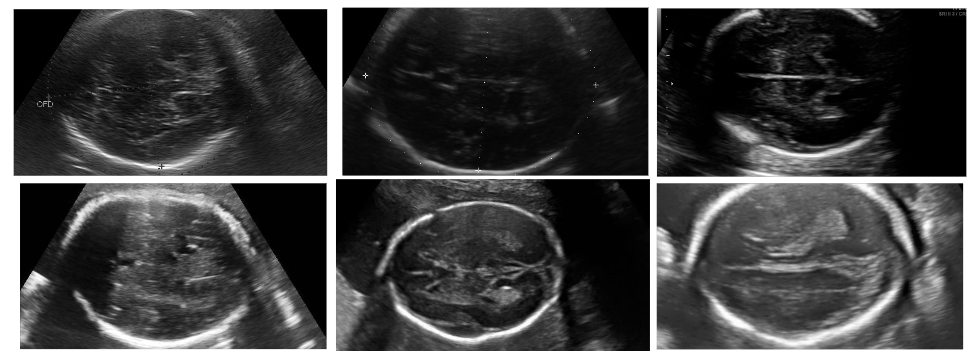}
\caption{Fetal standard planes for the brain acquired at different centers. From left to right and from top to bottom, the planes are from Spain, Ghana, Egypt, Malawi, Algeria, Uganda} 
\label{fig:sp_examples}
\end{figure}

All images are labeled with five standard plane categories: \textit{abdomen}, \textit{brain}, \textit{femur}, \textit{thorax} and \textit{other} (the latter includes other less common planes such as maternal cervix, umbilical cord or extremities). These labels were assigned by clinicians or technicians with extensive training in ultrasound fetal plane classification. Figure~\ref{fig:sp_examples} shows some examples of these images. The number of planes per category and country is shown in Table~\ref{tab:data}. There are two things to highlight about this distribution, the first one and the most evident is that the African datasets have a very few amount of samples in comparison with the Spanish dataset. In order to try to mitigate this unbalance we perform data augmentation to the African datasets by adding randomly rotated as well as randomly cropped images between a factor of 0.8 and 1.0 of its original size. The second point is that only the dataset from the Spanish hospitals include this last category, so we removed it from the study in order to avoid an even more unbalanced distribution. In this proof-of-concept study, we did not incorporate additional balancing techniques, as the primary goal was to test out the deployed platform rather than optimize or fine-tune the model's performance. Our focus was on demonstrating the feasibility of the proposal, leaving the exploration of advanced balancing methods for future research since it is true that in real life we will find unbalanced datasets specially when dealing with African rural health centers that could have less data

All of the images were preprocessed by resizing them to 256x256 pixels, allowing us to create mini-batches and still run given the compute limitations. Intensity values were normalized  between 0 and 1.

Since all datasets used in this study are publicly available, no ethical approval was needed to conduct the current study.


\begin{table}
\begin{center}
\caption{Number of images per category and country}
\label{tab:data}
\begin{tabular}{cccccc}
\toprule
Country & Abdomen & Brain & Femur & Thorax & Other\\
\midrule
Spain & 711 & 1781 & 1040 & 1718 & 4213 \\
Malawi & 25 & 25 & 25 & 25 & 0 \\
Egypt & 25 & 25 & 25 & 25 & 0 \\
Uganda & 25 & 25 & 25 & 0 & 0 \\
Ghana & 25 & 25 & 25 & 0 & 0 \\
Algeria & 25 & 25 & 25 & 25 & 0 \\
\bottomrule
\end{tabular}
\end{center}
\end{table}

\subsection{Training}
FL consists of a collaborative training in which the model weights are distributed among client nodes, each with its own local data. The models are distributed to and refined at various centers, then returned and aggregated on a central server. This process is repeated for several rounds, with the updated models being redistributed to the original nodes.  While there are several aggregation techniques available such as model selection, weighted averaging, median aggregation or voting aggregation, in this proof-of-concept study, we utilize federated averaging, which involves averaging the weights of all the different local models [16]. This aggregation process extracts insights from all datasets without transferring raw data beyond each client. 
Subsequently, the updated aggregated weights are broadcast back to all clients, who then continue training the aggregated model with their respective private datasets. This iterative process retains data locally and only shares model weights reducing privacy risks associated with data sharing. Additionally, it distributes computational load across clients.

In the current work, we implemented an FL platform using the Flower library (\emph{flwr}) [6], PyTorch [9] and PyTorch Lightning [10]. \emph{flwr} is used as a framework for communicating with the different nodes from the central server through Secure Socket Layer (SSL) protocol (guaranteeing a secure connection) and for performing the aggregation of the client models. Then, each client has its own local model that is trained with its local data, in this part, Lightning was used to obtain optimal speed and reliability by optimizing computation and memory usage, enabling more efficient model training. Additionally, we used the automatic mixed precision feature to accelerate training and reduce memory usage without sacrificing model accuracy. The optimizer used was SGD with a learning rate scheduler that dynamically adjusts the learning rate based on the training progress, avoiding unnecessary stagnation of the training loop.

Our framework includes different models and variants of them, but for this study we use ResNet-50 [11] since it provides a good trade-off between accuracy and resource usage.

In synchronous training, the primary challenge when using a Raspberry Pi in federated frameworks is that the training speed is limited by the slowest node. This is because the framework must wait for all clients to respond, potentially leading to impractically long training times. On the other hand, asynchronous training allows the server to proceed without waiting for all nodes to complete each round. However, this approach carries the risk that slower nodes, such as the Raspberry Pi, may not contribute sufficiently to the training. We decided to train using synchronous steps to have all the nodes contribute equally and to measure how much the whole system slows down when adding an edge device and thus, to determine the feasibility of using it.

To optimize performance across different nodes, we configured each node to train for 10 epochs per round, except for the Raspberry Pi, which was set to 3 epochs per round. We trained for 20 rounds, with each node having a training dataset size of $80\%$, except for the Raspberry Pi, which had a dataset size of only $40\%$. The testing, validation and training datasets were divided into non-overlapping partitions.

\subsection{Hardware}

A variety of hardware components was used in this study, with specific details provided in Table~\ref{tab:hardware}. Only one computer had access to a GPU for training, one node was the Raspberry Pi 5 with an ARM processor and the other computers trained only using the CPU.

\begin{table}
\caption{Hardware used for the different nodes during the training process.}
\label{tab:hardware}
\begin{tabular}{clccc}
\toprule
Node & Location &  Processor & GPU & OS  \\
\midrule
Central & Spain & Intel Core i7-9700K & RTX 2060 8GB
 & Ubuntu 22.04 LTS \\
1 & Argelia & Intel Core i7-8650U & - & Windows 11 Pro \\
2 & Ghana & Intel Core i5-1145G7 & - & Windows \\
3 & Spain & Arm Cortex-A76 & - & Ubuntu 24.04 Raspberry\\
4 & Spain & Intel Core i7-9700K & RTX 2080 8GB & Ubuntu 22.04 LTS\\
5 & Uganda & Intel 300  & - & Windows 10 Pro\\
6 & Uganda & Intel Core i7-6700 & - & Ubuntu 22.04 LTS\\
\bottomrule
\end{tabular}
\end{table}

\section{Results and discussion}

\begin{figure}
\centering
\includegraphics[width=\textwidth]{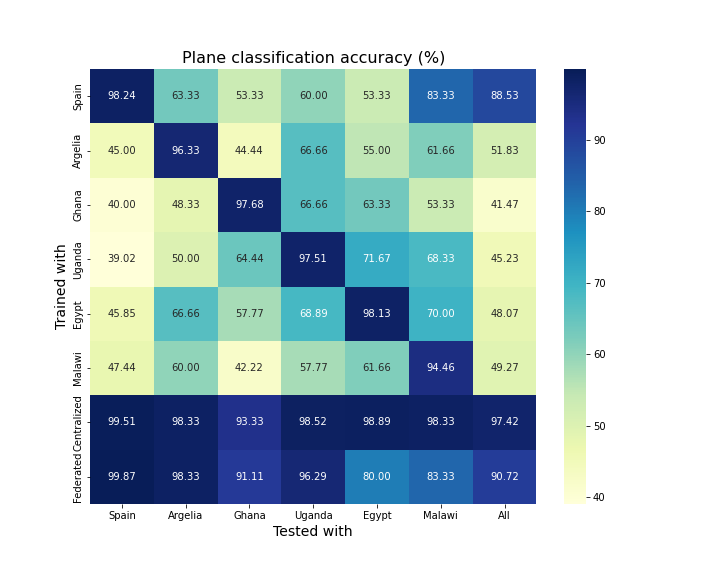}
\caption{Accuracy achieved with different datasets for Resnet-50}
\label{fig:heatmap}
\end{figure}

\begin{table}
\centering
\caption{Performance related information}
\label{tab:performance}
\begin{tabular}{ccccccc}
\toprule
Dataset & Device & Samples & Batch & Epochs/Round & Iter./s   & Avg Time/Round (min) \\
\midrule
Spain   & GPU       & 6549 & 8 & 10 & $\sim$ 15.6 & 10.48 \\
Malawi  & Raspberry & 240  & 2 & 3  & $\sim$ 0.3  & 26.78 \\
Egypt   & CPU       & 480  & 8 & 10 & $\sim$ 2.5  & 4.05 \\
Uganda  & CPU       & 360  & 8 & 10 & $\sim$ 0.57 & 12.96 \\
Ghana   & CPU       & 360  & 8 & 10 & $\sim$ 0.67 & 14.88 \\
Algeria & CPU       & 480  & 8 & 10 & $\sim$ 0.79 & 14.81 \\
\bottomrule
\end{tabular}
\end{table}

We trained 8 different models. Six local models were trained with the respective dataset from each node that represents each country. One centralized model with all the datasets merged together, that represents the situation in which all the centers manage and agree to transfer all their data to one single center, training node 4 in Table~\ref{tab:hardware}. And one final model trained with our federated framework, in which all the datasets were kept in their respective centers. Figure~\ref{fig:heatmap} shows the accuracy for each model when tested on the different datasets in order to measure the achieved generalizability.

\begin{figure}
\includegraphics[width=\textwidth]{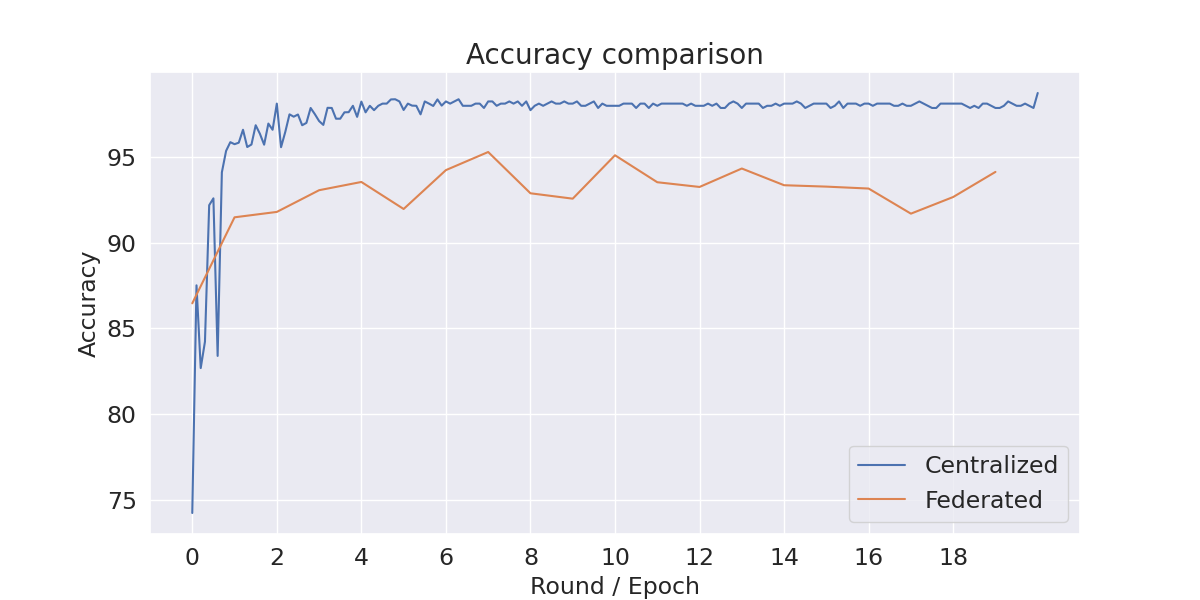}
\caption{Accuracy evolution during training of the centralized and federated models. }
\label{fig:accuracy}
\end{figure}

It is possible to observe that the models trained with individual datasets are not getting enough generalization when tested with external datasets. For example, the model trained with the Ugandan dataset achieves only $39\%$ of accuracy on the Spanish dataset. In contrast, the centralized model gets a high level of generalization across all test datasets. Lastly, our federated trained model also achieves a good level of generalization but without sharing data between the nodes and with important hardware constrains. Figure~\ref{fig:accuracy} shows the evolution of the accuracy during training process of centralized and federated approaches. While the centralized training is more stable and achieves higher performance, the accuracy of the federated model is comparable with very few rounds into the training.

The Raspberry Pi not only successfully processed and trained a model locally with its dataset but also could participate in the FL training among all of the other nodes. With this, we validate its capability as a low-cost, effective tool for AI deployment in resource-constrained settings. Also, we show the feasibility and potential of training models with a FL approach using edge devices such as Raspberry Pi.

Table~\ref{tab:performance} shows information related to the performance of every node. In all cases the epochs per round was fixed to 10 except in the case of the Raspberry, that was configured to perform only 3 to compensate for the performance gap with the other nodes. The batch size was also smaller in this case to avoid memory bottlenecks. By looking at the iterations per second and the average time per round, one can notice that the Raspberry Pi node is the slowest. However, the difference can be compensated by reducing the batch size and the epochs per round. While it is slowing the whole process, the accuracy is still high as we can corroborate in Figure~\ref{fig:heatmap}.

\section{Conclusions}

This study aims to be the beginning of a series of research efforts focused on utilizing FL in Africa. We demonstrated that using limited resources devices within a federated project such as Raspberry Pi is feasible. Our developed platform was successfully deployed and tested in a realistic scenario with diverse constrained devices. This deployment marks a significant starting point in our project, demonstrating both the robustness of our solution and that such a framework will be able to handle future real-world applications. The platform is now fully operational and ready to deliver its intended benefits to users encouraging the formation of partnerships between local healthcare providers, international research institutions, and technology developers.

\begin{credits}
\subsubsection{\ackname}
 This research was funded by a grant from the European Research Council (ERC) under the European Union’s Horizon Europe research and innovation programme (AIMIX project - Grant Agreement No. 101044779).

\end{credits}
%
\section{Bibliography}

[1] Martin-Isla, C., Campello, V. M., Izquierdo, C., Raisi-Estabragh, Z., Baeßler, B., Petersen, S. E., \& Lekadir, K. (2020). Image-based cardiac diagnosis with machine learning: A review. Frontiers in Cardiovascular Medicine, 7, 1.

[2] Sohan, M. F., \& Basalamah, A. (2023). A systematic review on federated learning in medical image analysis. IEEE Access, 11, 28628-28644. 

[3] Linardos, A., Kushibar, K., Walsh, S., Gkontra, P., \& Lekadir, K. (2022). Federated learning for multi-center imaging diagnostics: A simulation study in cardiovascular disease. Scientific Reports, 12(1), 3551.

[4] Rehman, M. H., Hugo Lopez Pinaya, W., Nachev, P., Teo, J. T., Ourselin, S., \& Cardoso, M. J. (2023, October). Federated learning for medical imaging radiology. British Journal of Radiology, 96(1150), 20220890. 

[5] Sendra-Balcells, C., Campello, V. M., Torrents-Barrena, J., Ahmed, Y. A., Elattar, M., Ohene-Botwe, B., Nyangulu, P., Stones, W., Ammar, M., Benamer, L. N., \& others. (2023). Generalisability of fetal ultrasound deep learning models to low-resource imaging settings in five African countries. Scientific Reports, 13(1), 2728.

[6] Beutel, D. J., Topal, T., Mathur, A., Qiu, X., Fernandez-Marques, J., Gao, Y., Sani, L., Kwing, H. L., Parcollet, T., Gusmão, P. P. B., \& Lane, N. D. (2020). Flower: A friendly federated learning research framework. arXiv preprint arXiv:2007.14390.

[7] Burgos-Artizzu, X. P., Coronado-Gutiérrez, D., Valenzuela-Alcaraz, B., Bonet-Carne, E., Eixarch, E., Crispi, F., \& Gratacós, E. (2020). Evaluation of deep convolutional neural networks for automatic classification of common maternal fetal ultrasound planes. Scientific Reports, 10(1), 10200.

[8] Soltan, A. A. S., Thakur, A., Yang, J., Chauhan, A., D'Cruz, L. G., Dickson, P., Soltan, M. A., Thickett, D. R., Eyre, D. W., Zhu, T., \& Clifton, D. A. (2024, February). A scalable federated learning solution for secondary care using low-cost microcomputing: Privacy-preserving development and evaluation of a COVID-19 screening test in UK hospitals. The Lancet Digital Health, 6(2), e93-e104. 

[9] Ansel, J., Yang, E., He, H., Gimelshein, N., Jain, A., Voznesensky, M., Bao, B., Bell, P., Berard, D., Burovski, E., Chauhan, G., Chourdia, A., Constable, W., Desmaison, A., DeVito, Z., Ellison, E., Feng, W., Gong, J., Gschwind, M., ... Chintala, S. (2024, April). PyTorch 2: Faster machine learning through dynamic python bytecode transformation and graph compilation. In Proceedings of the 29th ACM International Conference on Architectural Support for Programming Languages and Operating Systems, Volume 2 (ASPLOS '24). ACM. 

[10] Falcon, W., \& The PyTorch Lightning team. (2019, March). PyTorch Lightning (Version 1.4) [Computer software]. https://github.com/Lightning-AI/lightning

[11] He, K., Zhang, X., Ren, S., \& Sun, J. (2016). Deep residual learning for image recognition. In Proceedings of the IEEE conference on computer vision and pattern recognition (pp. 770-778).

[12] Moshawrab, M., Adda, M., Bouzouane, A., Ibrahim, H., \& Raad, A. (2023). Reviewing federated learning aggregation algorithms; strategies, contributions, limitations and future perspectives. Electronics, 12(10), 2287. 

[13] Mollura, D. J., Culp, M. P., Pollack, E., Battino, G., Scheel, J. R., Mango, V. L., Elahi, A., Schweitzer, A., \& Dako, F. (2020, December). Artificial intelligence in low- and middle-income countries: Innovating global health radiology. Radiology, 297(3), 513-520. https://doi.org/10.1148/radiol.2020201434

[14] Birur, N. P., Song, B., Sunny, S. P., Keerthi, G., Mendonca, P., Mukhia, N., Li, S., Patrick, S., Shubha, G., Subhashini, A. R., Imchen, T., Leivon, S. T., Kolur, T., Shetty, V., Vidya Bhushan, R., Vaibhavi, D., Rajeev, S., Pednekar, S., Banik, A. D., ... Kuriakose, M. A. (2022, August). Field validation of deep learning based point-of-care device for early detection of oral malignant and potentially malignant disorders. Scientific Reports, 12(1), 14283. 

[15] Maraci, M. A., Yaqub, M., Craik, R., Beriwal, S., Self, A., von Dadelszen, P., Papageorghiou, A., \& Noble, J. A. (2020, January). Toward point-of-care ultrasound estimation of fetal gestational age from the trans-cerebellar diameter using CNN-based ultrasound image analysis. Journal of Medical Imaging, 7(1), 014501. 

[16] McMahan, B., Moore, E., Ramage, D., Hampson, S., \& y Arcas, B. A. (2017). Communication-efficient learning of deep networks from decentralized data. In Proceedings of Artificial Intelligence and Statistics (pp. 1273-1282). PMLR.

[17] Yang, Q., Liu, Y., Chen, T., \& Tong, Y. (2019). Federated machine learning: Concept and applications. ACM Transactions on Intelligent Systems and Technology (TIST), 10(2), 1-19.

[18] Rieke, N., Hancox, J., Li, W., Milletari, F., Roth, H. R., Albarqouni, S., Bakas, S., Galtier, M. N., Landman, B. A., Maier-Hein, K., ... \& others. (2020). The future of digital health with federated learning. NPJ Digital Medicine, 3(1), 1-7.

[19] Dayan, I., Roth, H. R., Zhong, A., Harouni, A., Gentili, A., Abidin, A. Z., Liu, A., Beardsworth Costa, A., Wood, B. J., Tsai, C. S., ... \& others. (2021). Federated learning for medical imaging: Our experience and next steps. Journal of the American College of Radiology, 18(9), 1213-1221.

\bibliography{bibliography}

\end{document}